\documentclass[11pt]{article}
\usepackage[final]{acl}

\usepackage{times}
\usepackage{latexsym}

\usepackage[T1]{fontenc}

\usepackage[utf8]{inputenc}
\usepackage{dashrule}
\usepackage{microtype}
\usepackage[table,xcdraw]{xcolor}  
\usepackage{inconsolata}

\usepackage{times}
\usepackage{latexsym}
\usepackage[edges,linguistics]{forest}
\usepackage{textcomp}
\usepackage{amsfonts}
\usepackage{amssymb}
\usepackage{amsmath}
\usepackage{xcolor}
\usepackage{tabularx}
\usepackage{booktabs}

\usepackage{booktabs}    
\usepackage{multirow}    
\usepackage{makecell}    
\usepackage{graphicx}    
\usepackage{adjustbox}
\usepackage[normalem]{ulem}
\usepackage{listings}
\usepackage{xcolor}
\usepackage[most]{tcolorbox} 
\definecolor{codegreen}{rgb}{0,0.6,0}
\definecolor{codegray}{rgb}{0.5,0.5,0.5}
\definecolor{codepurple}{rgb}{0.58,0,0.82}
\definecolor{backcolour}{rgb}{0.98,0.98,0.98} 

\definecolor{annotred}{RGB}{200, 40, 40}  
\definecolor{bgmain}{RGB}{248, 249, 250}  
\definecolor{bgtool}{RGB}{255, 255, 255}  

\lstdefinelanguage{json}{
    basicstyle=\ttfamily\scriptsize, 
    backgroundcolor=\color{backcolour},   
    commentstyle=\color{codegreen},
    keywordstyle=\color{magenta},
    stringstyle=\color{codepurple},
    columns=fullflexible, 
    showstringspaces=false,
    keepspaces=true, 
    breaklines=true,
    breakatwhitespace=true, 
    frame=none, 
    literate=
     *{:}{{{\color{black}{:}}}}{1}
      {,}{{{\color{black}{,}}}}{1}
      {\{}{{{\color{black}{\{}}}}{1}
      {\}}{{{\color{black}{\}}}}}{1}
      {[}{{{\color{black}{[}}}}{1}
      {]}{{{\color{black}{]}}}}{1},
}
\usepackage{enumitem}
\usepackage{minted}
\usepackage[most]{tcolorbox}
\newtcolorbox{paperbox}[2][]{%
  enhanced,
  colback=white,
  colframe=black,
  fonttitle=\bfseries,
  title={#2},
  #1
}
\usepackage{pifont}   
\usepackage{xcolor}   
\definecolor{myblue}{RGB}{0,114,189}
\definecolor{darkgreen}{RGB}{0,100,0}  
\definecolor{darkred}{RGB}{139,0,0}    
  
\title{\textsc{DeepMed}: Building a Medical DeepResearch Agent via Multi-hop Med-Search Data and Turn-Controlled Agentic Training \& Inference}
\author{
  \textbf{Zihan Wang\textsuperscript{\(\spadesuit,\heartsuit\)}}\thanks{Equal contribution.}%
  \quad \textbf{Hao Wang\textsuperscript{\(\heartsuit\)}}\footnotemark[1]%
  \quad \textbf{Shi Feng\textsuperscript{\(\spadesuit\)}}\thanks{Corresponding author.}%
  \quad \textbf{Xiaocui Yang\textsuperscript{\(\spadesuit\)}}%
  \quad \textbf{Daling Wang\textsuperscript{\(\spadesuit\)}}\\
  \textbf{Yiqun Zhang\textsuperscript{\(\spadesuit\)}}%
  \quad \textbf{Jinghao Lin\textsuperscript{\(\heartsuit\)}}%
  \quad \textbf{Haihua Yang\textsuperscript{\(\heartsuit\)}}%
  \quad \textbf{Xiaozhong Ji\textsuperscript{\(\heartsuit\)}}\\
  \textsuperscript{\(\spadesuit\)}Northeastern University,  
  \textsuperscript{\(\heartsuit\)}ByteDance\\
  \{2310744@stu.neu.edu.cn, fengshi@cse.neu.edu.cn\}
}

\begin{document}
\maketitle
\begin{abstract}
Medical reasoning models remain constrained by parametric knowledge and are thus susceptible to forgetting and hallucinations. 
DeepResearch (DR) models ground outputs in verifiable evidence from tools and perform strongly in general domains, but their direct transfer to medical field yields relatively limited gains. 
We attribute this to two gaps: task characteristic and tool-use scaling. 
Medical questions require evidence interpretation in a knowledge-intensive clinical context; while general DR models can retrieve information, they often lack clinical-context reasoning and thus “find it but fail to use it,” leaving performance limited by medical abilities. 
Moreover, in medical scenarios, blindly scaling tool-call can inject noisy context, derailing sensitive medical reasoning and prompting repetitive evidence-seeking along incorrect paths.
Therefore, we propose \textsc{DeepMed}. 
For data, we deploy a multi-hop med-search QA synthesis method supporting the model to apply the DR paradigm in medical contexts. 
For training, we introduce a difficulty-aware turn-penalty to suppress excessive tool-call growth. 
For inference, we bring a monitor to help validate hypotheses within a controlled number of steps and avoid context rot. 
Overall, on seven medical benchmarks, \textsc{DeepMed} improves its base model by 9.79\% on average and outperforms larger medical reasoning and DR models.

\end{abstract}

\section{Introduction}
In the medical field, reasoning models~\cite{liu2025beyond,zhang2025med,huang2025m1} can achieve stronger diagnostic performance than conventional LLMs, while they remain constrained by the limits of parametric knowledge~\cite{chen2024huatuogpto1medicalcomplexreasoning}. 
It can induce hallucinations and spurious attributions, leading to diagnostic bias and potential clinical risks~\cite{feng2025iprar,thirunavukarasu2023llmmedicine}. 
While the emerging DeepResearch (DR) paradigm in general domains tightly integrates web search with model reasoning, grounding generation in verifiable external evidence and thereby reducing hallucinations~\cite{li2025websailor2, li2025websailor1}. 
DR has shown strong capability in complex problem solving and multi-step reasoning, particularly in open-domain question answering (QA)~\cite{wei2025browsecomp,zhou2025browsecomp}. 
Nevertheless, directly transferring general-purpose DR models to medical settings remains highly challenging. 
\begin{figure}[t!]            
  \centering
  \includegraphics[width=0.98\linewidth]{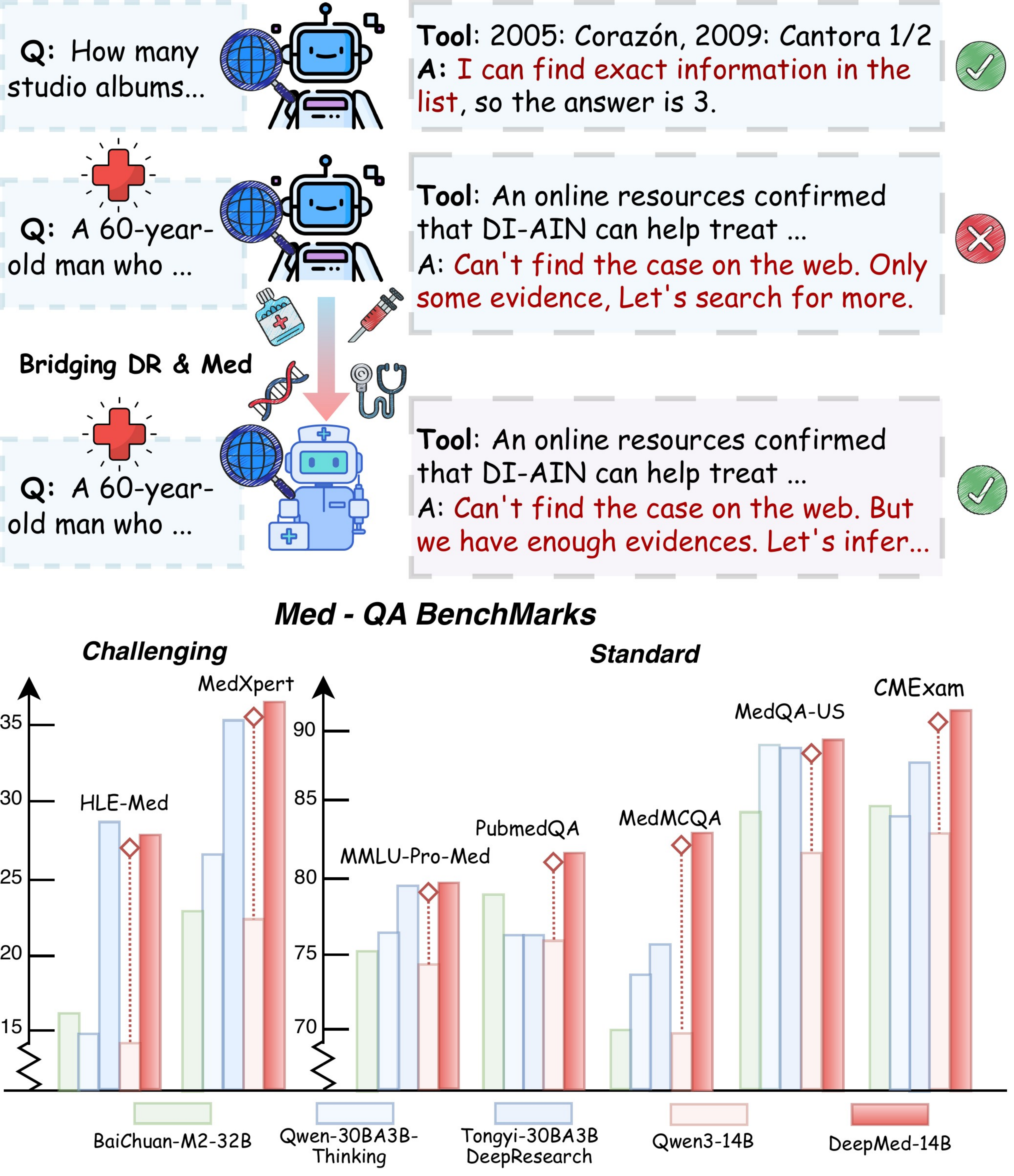}
  \caption{General search tasks typically provide definitive answers online, but medical tasks are different; while online information can assist, the final answer relies on medical reasoning \textbf{(up)}.
  \textsc{DeepMed} shows significant improvements over its base reasoning model on medical benchmarks, and outperforms larger medical reasoning model and general DR models \textbf{(down)}.}
  \label{fig:intro}
  \vspace{-1.5em}
\end{figure}

We evaluated Tongyi-DeepResearch~\cite{team2025tongyi}, a strong DR model trained with large-scale agentic training, on medical benchmarks and found that its gains in medical tasks are modest against its base reasoning model - Qwen3-30BA3B-Thinking as shown in Fig.~\ref{fig:intro}, especially when contrasted with its substantial improvements on general search tasks. 
Two key gaps are identified in translating DR capabilities to medical diagnosis. 
\textbf{(i) Task characteristic mismatch.} General DR models training primarily targets general factual queries (e.g., dates, people, and entities) ~\cite{wei2025browsecomp,zhou2025browsecomp}. 
In contrast, medical questions require differential reasoning and decision making in a knowledge-intensive clinical context: the model must not only retrieve information, but also interpret its clinical implications, weigh the strength and conflicts of evidence, and use medical priors to filter key cues. 
\textbf{(ii) Tool-scaling profit mismatch.} 
During both training and inference, general DR systems often improve performance by scaling the number of tool calls to gather more information ~\cite{li2025websailor2,li2025websailor1} like Fig.~\ref{fig:intro}. 
In medical diagnosis, however, more is not necessarily better. 
Excessive searching introduces noise that accelerates context rot~\cite{hong2025contextrot}, degrading the context—an effect especially pronounced in medicine due to the sensitivity and error-proneness of medical information~\cite{neha2025retrieval,ke2025retrieval}, which ultimately constrains model performance. 
Furthermore, injecting large medical context can trigger an “over-evidence” phenomenon: where the model struggles to integrate the accumulated information and thus repeatedly calls tools to re-verify what it already knows instead of advancing its reasoning, creating a self-reinforcing negative feedback loop. 

To migrate these gaps, we propose \textsc{DeepMed}, a DeepResearch model tailored for medical scenarios. 
We introduce a web-based method to synthesize multi-hop medical questions used for agentic Supervised Fine-Tuning (SFT) stage, it strengthen multi-hop search ability in the medical knowledge-rich context. 
During agentic Reinforment Learning (RL), we train the model using challenging diagnostic tasks to further improve its ability to integrate search and medical reasoning capabilities. 
To control the unrestrained tool scaling often found in Agentic RL~\cite{feng2025iprar}, we deploy a difficulty-aware turn-penalty to implicitly suppress it while training. 
For easier samples, the model is encouraged to make fewer tool calls to reduce redundant exploration; for harder samples, it is allowed a moderate increase in tool usage to acquire crucial evidence. 
For inference, we deploy a monitor to explicitly rescue the model from ``over-evidence''. 
When the model begins to repeatedly validate a candidate answer, it halt further tool exploration in time and help it to finalize the response. 
Ultimately, \textsc{DeepMed} achieves substantial average improvements over the base model, Qwen3-14B, as shown in Fig.~\ref{fig:intro}. 

Our contributions can be summarized as follows: 
\begin{itemize}[noitemsep, left=0pt]
\item We identify two gaps that arise when applying DR paradigm to medical field—one in task characteristics and the other in the benefits of tool scaling—and address them with \textsc{DeepMed}.
\item \textsc{DeepMed} improves by an average of \textbf{13.92\%} on the two challenging benchmarks and \textbf{8.13\%} on the five standard benchmarks against its base model and outperforms several medical reasoning models and DR  models with more parameters or trained on more data. 
\item We validate that \textsc{DeepMed} can leverage web medical evidence to reduce hallucinations and self-correct in reasoning, demonstrating the promise of DeepResearch in the medical field. 
\end{itemize}


\section{Related Work}
\subsection{DeepResearch Models}
DeepResearch models utilize web search tools and real-world network knowledge to find or verify the information needed for inference~\cite{zhang2025deep}. 
OpenAI DeepResearch is the most well-known example, and this capability has become a target for improvement for various research institutions and companies~\cite{zeng2025glm,team2025kimi}. 
Early works such as Webwalker/Search-R1 ~\cite{wu2025webwalker,jin2025search} explore the capabilities of this paradigm, achieving significant improvements on some search benches. Subsequent works such as Websailor/Deepdive~\cite{lu2025deepdive,li2025websailor1} further validate that multi-hop QA problems and agentic RL are key to unlocking these capabilities. 
This paradigm exhibits strong generalization, bringing improvements even to extremely hard QA problems such as Humanity-Last-Exam (HLE)~\cite{phan2025humanity}. 
MedResearcher-R1~\cite{yu2025medresearcher} first applies this capability to MedBrowseComp~\cite{chen2025medbrowsecomp}, a medical search benchmark that does not target diagnostic or clinical decision-making performance. 
Accordingly, we stress that the central challenge is bridging search and medicne: medical systems must go beyond better retrieval to use retrieved evidence to drive and verify diagnostic hypotheses, enhancing accuracy while eliminating illusory reasoning. 

\subsection{Medical Reasoning Models}
LLMs are widely used in medicine~\citep{liu2024surveymedicallargelanguage, wu2024pmc, qiu2024buildingmultilinguallanguagemodel}, and stronger chain-of-thought reasoning often brings substantial performance gains. 
HuatuoGPT-o1~\citep{chen2024huatuogpto1medicalcomplexreasoning} trains a 70B model with SFT followed by RL and achieves better performance on medical reasoning benchmarks. 
In parallel, M1~\cite{huang2025m1}, AlphaMed~\cite{liu2025beyond}, and BioMed-R1~\citep{thapa2025disentanglingreasoningknowledgemedical} focus on pushing the upper bound of medical reasoning via reinforcement learning. Industrial efforts such as Baichuan-M2~\cite{dou2025baichuan} and Quark-Med~\cite{li2025quarkmed} further demonstrate the promise of medical reasoning models. 
Despite these advances, most methods still rely on internal reasoning, making them prone to forgetting and hallucinations. 
Combining medical reasoning with external knowledge and tool use is therefore key for robustness and real-world utility. While RAG-style methods help~\cite{xiong-etal-2024-benchmarking,zhao2025medrag}, their dependence on private databases and loose coupling between reasoning and dynamically retrieved evidence limit multi-hop solving~\cite{jin2025search}. In contrast, the open web offers abundant reliable medical information, highlighting the promise of DR models for medicine. 

\section{Preliminaries}
\paragraph{DeepResearch Paradigm}
Standard “thinking” models implement slow thinking by first producing a chain of thought (CoT)~\cite{jaech2024openai,guo2025deepseek} wrapped in \texttt{<think>} tags to analyze the input problem $x$, and then output the final answer $y$. The resulting trace can be written as $\mathcal{T}=\langle x, (\mathrm{CoT}), y \rangle$.
In contrast, DeepResearch models interleave tool use with the reasoning process, typically following the ReAct~\cite{yao2022react} paradigm. After \texttt{<think>}, tool invocations are wrapped in \texttt{<tool\_call>} tags. A downstream tool server executes the call and returns the result wrapped in \texttt{<tool\_response>}, which is then appended to the model context. This yields a multi-turn interaction, and the model outputs the final answer only after it stops calling tools.
\begin{equation}
\label{equ:tool_trace}
\begin{aligned}
\mathcal{T}
= \langle x,&\underbrace{(\mathrm{CoT}_1, \mathrm{Tool}^{\mathrm{call}}_1, \mathrm{Tool}^{\mathrm{res}}_1)}_{\text{Turn } 1},\dots, \\ 
&\underbrace{(\mathrm{CoT}_N, \mathrm{Tool}^{\mathrm{call}}_N, \mathrm{Tool}^{\mathrm{res}}_N)}_{\text{Turn } N}, y \rangle ,
\end{aligned}
\end{equation}

\paragraph{Tool Configuration} 
We leverage web-based external knowledge—preferred for its broad coverage and accessibility—as the primary grounding source for \textsc{DeepMed}. 
Specifically, the model accesses online information through two tools, \texttt{Search} and \texttt{Visit}. \texttt{Search} takes a \texttt{query} as input and returns a list of relevant URLs along with brief snippets. \texttt{Visit} takes a target \texttt{Url} and a \texttt{Goal} as input, retrieves the full webpage content, and then returns a condensed summary of the information that is most relevant to the specified goal~\cite{team2025tongyi}. 

\section{\textsc{DeepMed} Methodology}
In this section, we formalize the key components of \textsc{DeepMed} and present the underlying methodology with equations. 
Our training procedure consists of two stages: (i) Warm-up Agentic SFT (ASFT) with Multi-hop Med-Search QA, and
(ii) Agentic RL (ARL) with Hard Med-Diagnosis QA.
The first stage teaches the model to invoke tools properly in a medically rich context, while the second stage further couples retrieval with medical reasoning, unlocking the model's potential on medical tasks.
For inference, we introduce (iii) Inference with Over-Evidence Monitor, which prevents the model from falling into the over-evidence trap. 
\begin{figure*}[t!]
  \centering
  \resizebox{\textwidth}{!}{%
    \includegraphics{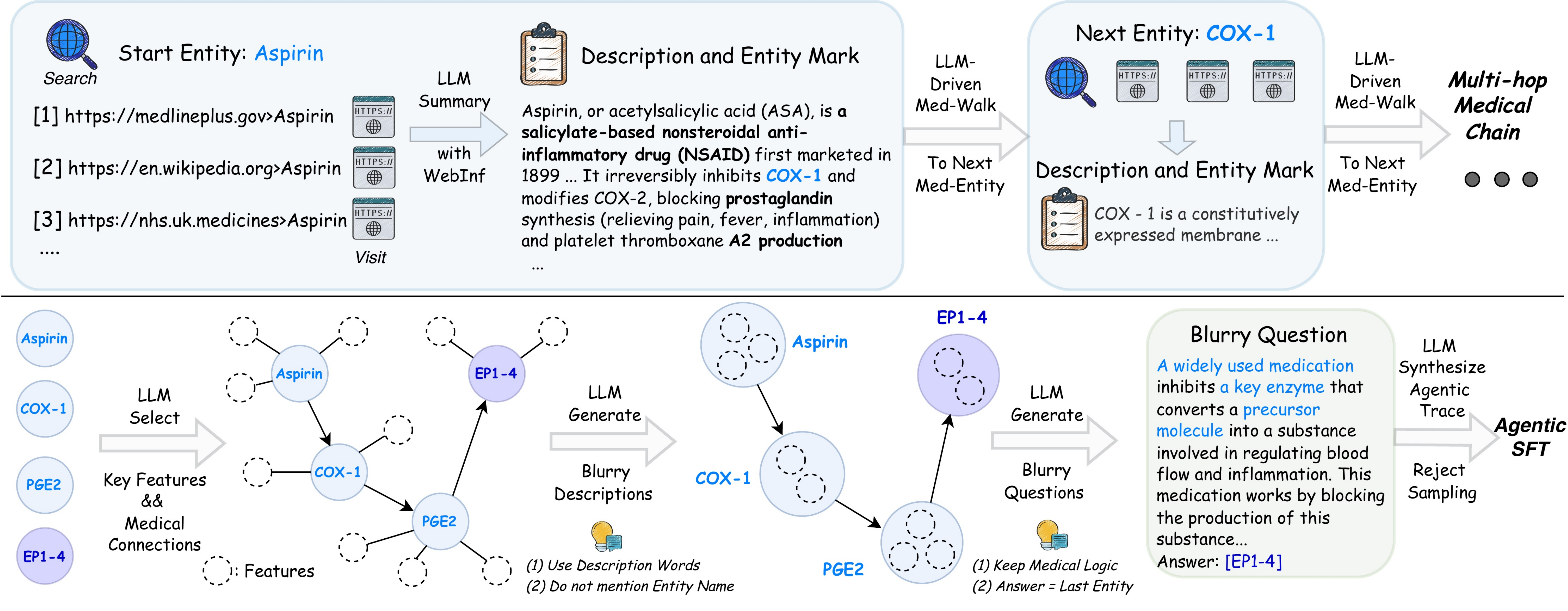}%
  }
  \caption{The multi-hop Med-Search data synthesis workflow for Agentic SFT of \textsc{DeepMed}. 
  The top panel illustrates how we synthesize multi-hop medical chains that emphasize logical relations by leveraging multi-source web evidence. 
  The bottom panel shows how these chains are transformed into Med-Search QA.}
  \label{fig:data_syn}
  \vspace{-1em}
\end{figure*}

\subsection{ASFT with Multi-hop Med-Search QA}
Agentic SFT is a critical stage in training DeepResearch models. 
It serves to warm-up the model, reshape its reasoning paradigm, and teach it to interface effectively with external knowledge sources by tools. 
Prior work has consistently shown that skipping this stage typically results in degraded performance~\cite{team2025tongyi,wu2025webwalker}. 
\subsubsection{Multi-hop Med-Search QA Synthesis}
\paragraph{Multi-hop Medical Chain Construction}
The essence of multi-hop QA lies in constructing an entity chain that encourages the model to decompose a question step by step along a logical path before arriving at the final answer. 
In this work, we synthesize such medical chains using web-based knowledge as shown in Fig.~\ref{fig:data_syn}. 
Compared with knowledge-graph-based approaches \cite{lu2025deepdive,li2025websailor1}, our web-based method is more flexible and scalable, and it can leverage diverse online sources to build more reliable medical links. 
Concretely, we start from a medical keyword or entity (e.g., a disease or a drug). 
We collect information about the entity from authoritative websites, extract relevant medical facts, and consolidate them into a short entity summary. 
Then identify candidate related nodes from the summary and, guide by medical logic (e.g., key disease characteristics, indicated treatments, or a drug's key ingredients and targets), select the most relevant entity as the next hop. 
Repeating this process forms a multi-hop sequence, where each hop is connected by verifiable medical relations. 
By iteratively traversing these links, we can flexibly extend the number of hops and ultimately construct a multi-hop medical chain from a start entity to an end entity. 

\paragraph{Med-Search QA Generation}
After constructing the medical chain, we translate its underlying logical relations into a multi-hop QA problem that requires step-by-step reasoning. 
A central design principle is entity obfuscation (identity blurring): we prohibit explicitly mentioning any major entity names in the question. 
As a result, the model must rely on tool use and medical reasoning to infer entity identities from functional roles and interactions, recover the latent chain, and ultimately derive the answer. 
To achieve this, we use an LLM (e.g., Gemini-2.5-pro) to obfuscate each entity description by paraphrasing it into an anonymized form that preserves only key distinguishing attributes. 
We then prompt the LLM to compose these obfuscated descriptions into a coherent Med-Search QA that faithfully reflects the original logical chain. 

\paragraph{Quality and Difficulty Control} 
For quality control, we use strong models (e.g., GPT-5) to filter the synthesized questions using full contextual information, focusing primarily on validating the soundness of the underlying logical chains. To increase difficulty, we deliberately extend the chain length (i.e., the number of hops), making the reasoning more complex and involving more entities. After question synthesis, we apply a pass@4 filter using Qwen-14B with a single tool call. We retain only questions that the model answers correctly at most once (i.e., 0/4 or 1/4), ensuring that the selected questions cannot be reliably solved by internal reasoning alone or by a trivial single-step retrieval.
A case question is shown in Appendix.~\ref{appendix:data}. 
\begin{figure*}[t!]
  \centering
  \resizebox{\textwidth}{!}{%
    \includegraphics{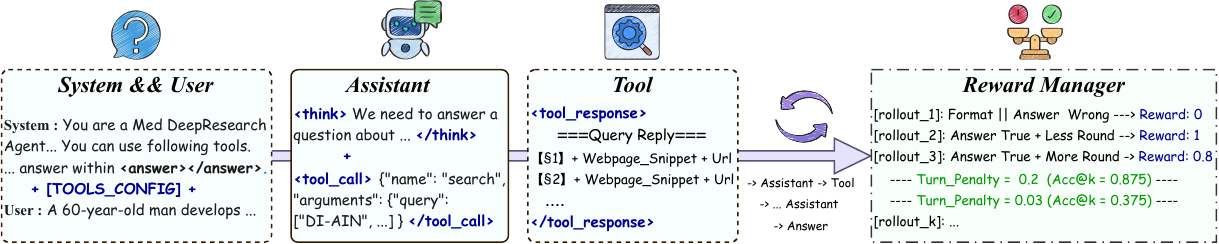}%
  }
  \caption{A full rollout and its reward evaluation are shown. Only segments inside the solid boxes contribute to the loss. Incorrect format or answer receives zero reward, and even correct rollouts are penalized for excessive rounds. }
  \label{fig:reward}
  \vspace{-1em}
\end{figure*}
\subsubsection{Warm-up ASFT}
Tools are wrapped as an MCP server and paired with an existing model to synthesize solution trajectories. Rejection sampling is then applied to retain only correct trajectories with a reasonable number of tool-use rounds. The resulting trajectories follow the format in Eq.~\ref{equ:tool_trace}. For training, a multi-turn dialogue-style formulation is used: the loss on non-model-generated segments (i.e., $\mathrm{Tool}^{\mathrm{res}}$) is masked out, and the standard SFT objective is applied to all remaining tokens, as defined in Eq.~\ref{equ:sft}. 
\begin{equation}
\label{equ:sft}
\mathcal{L}_{\mathrm{SFT}}(\theta)
= - \sum_{t=1}^{|\mathcal{T}|} M_t \cdot \log \pi_\theta\!\left(a_t \mid a_{<t}, x\right),
\end{equation}
$a$ denotes the tokens of the trajectory and $M_t \in \{0, 1\}$ is a binary mask. 
Specifically, $M_1 = 0$ if token $a_t$ falls within $x$ or $\mathrm{Tool}^{\mathrm{res}}$, and $M_t = 1$ for tokens within $\mathrm{CoT}$ and $\mathrm{Tool}^{\mathrm{Call}}$.

\subsection{ARL with Hard Medical QA}
ASFT teaches the model to use tools effectively in medical settings, delivering substantial gains over the base model. 
Building on this, we apply Agentic RL (ARL) on diagnosis-oriented problems to further strengthen medical search–reasoning integration and performance. 
Fig.~\ref{fig:reward} shows the workflow. 
\subsubsection{Challenging Medical QA Filter}
Following prior work~\cite{liu2025beyond,huang2025m1}, we curate a hard subset from the training split of existing datasets (e.g., MedQA-USMLE). 
The filtering proceeds in two stages: 
(i) we run Qwen-14B with direct (tool-free) reasoning for four trials and discard any questions with accuracy strictly above 50\%. 
For remaining questions, (ii) we run Qwen-14B with a simple tool-assisted setup (typically a single tool call) for four trials, and keep only those with accuracy 0 or 0.25. 

\subsubsection{ARL with Difficult-aware Turn-Penalty}
We utilize GRPO~\cite{shao2024deepseekmath} as the RL algorithm. 
As in ASFT, only the model-generated components (i.e., $\mathrm{CoT}$ and $\mathrm{Tool}^{\mathrm{call}}$) participate in optimization in ARL. 
For each medical query $q$, we sample a group of $G$ outputs from the old policy. 
Following prior work~\cite{lu2025deepdive,yu2025dapo}, we remove the KL-divergence regularization term. 
The policy is updated by maximizing the objective as Equ.~\ref{equ:grpo}. 
\begin{equation}
\label{equ:grpo}
\begin{aligned}
    \mathcal{L}_{\mathrm{GRPO}}(\theta)=\frac{1}{G} \sum_{i=1}^G \min \big(  \rho_i \hat{A}_i, 
    \text{clip}(\cdot)\hat{A}_i \big)
\end{aligned}
\end{equation}

\noindent where $\rho_i$ is the importance sampling ratio, and $clip$ is the clip range. 
The advantage $\hat{A}_i$ is computed by normalizing the rewards within the group average:
\begin{equation}
    \hat{A}_i = \frac{r_i - \text{mean}(r)}{std(r)}
\end{equation}
\noindent where $r_i$ is the total reward for trajectory $y_i$.

\paragraph{Reward Design}
In standard GRPO, rewards depend only on answer correctness, whereas in agentic training, format error (e.g., invalid tool calls, over long) are also treated as incorrect~\cite{team2025tongyi}. 
Meanwhile, we observe an ``over-evidence'' behavior, where the model keeps invoking tools redundantly despite already being able to answer, which yields low-quality rollouts and hinders policy learning. 
To address this, we introduce a turn-penalty for successful but overly long rollouts. 
Our overall reward is defined as Equ.~\ref{equ:reward_piecewise}
\begin{equation}
\label{equ:reward_piecewise}
r_i=
\begin{cases}
1, & F \wedge A \wedge T\\[2pt]
1-\lambda\, r^{\text{turn}}_i, & F \wedge A \wedge \neg T\\[2pt]
0, & \neg F \ \ \text{or}\ \ \neg A
\end{cases}
\end{equation}
where $(F,A)$ indicates whether the trajectory passes the format and answer checks, and $T$ is the turn-budget check defined as $T \triangleq \big(N_i \le \bar{N}^G_{suc}\big)$. $N_i$ is the number of interaction rounds and $\bar{N}^G_{\text{suc
}}$ is the mean turn count over true rollouts in the group $G$. 
$\lambda$ controls the penalty strength. 
The turn-penalty is defined as: 
\begin{equation}
\label{equ:rturn_piecewise}
r^{\text{turn}}_i= \omega f\big(N_i - \bar{N}^G_{\text{suc}}\big)
\end{equation}
Here, $f(\cdot)$ denotes a smoothing function; in our implementation, we use the logarithm. 
The coefficient $\omega$ is difficulty-adaptive: we estimate problem difficulty using the model's average rollout accuracy and assign smaller $\omega$ to harder questions. 

\begin{table*}[!t]
\centering
\small
\resizebox{\textwidth}{!}{%
\begin{tabular}{lccccccc}
\toprule
\toprule
Models/Datasets & HLE-Med & MedXpert & MMLU-Pro-Med & PubMedQA & MedMCQA & MedQA-USMLE & CMExam \\
\midrule
\multicolumn{8}{c}{\textbf{Awesome General Models}} \\
\midrule
\addlinespace[0.25em]
Gemini2.5-Pro & 14.36 & 39.27 & 84.30 & 76.00 & 84.13 & 92.22 & 87.40 \\
Kimi-K2-Thinking-1TB~\cite{team2025kimi} & 18.79 & 41.59 & 82.74 & 76.60 & 80.30 & 94.34 & 88.47 \\
Deepseek-v3.2-685B~\cite{liu2025deepseek} & 19.46 & 43.84 & 83.19 & 75.80 & 81.38 & 93.72 & 91.84 \\
\rowcolor{gray!13} Qwen3-30BA3B-Thinking~\cite{yang2025qwen3} & 14.77 & 26.40 & 76.94 & 76.00 & 73.11 & 88.14 & 83.81 \\
\rowcolor{gray!13} Qwen3-14B~\cite{yang2025qwen3} & 12.75 & 22.40 & 74.53 & 75.40 & 68.18 & 82.17 & 82.79 \\
\midrule
\multicolumn{8}{c}{\textbf{Medical Reasoning Models}} \\
\midrule
\addlinespace[0.25em]
HuatuoGPT-o1-70B~\cite{chen2024huatuogpto1medicalcomplexreasoning} & 11.60 & 26.50 & 76.61 & 80.60 & 73.61 & 83.27 & 70.42 \\
M1-1K-7B~\cite{huang2025m1} & 14.77 & 16.70 & 65.73 & 77.50 & 58.26 & 71.56 & 70.64 \\
M1-1K-32B~\cite{huang2025m1} & 16.78 & 25.20 & 71.53 & 77.60 & 67.37 & 83.50 & 79.03 \\
MedResaon-8B~\cite{wu2025medreason} & 22.40 & 16.40 & 63.13 & 77.60 & 57.25 & 68.40 & 68.95 \\
AlphaMed-70B~\cite{liu2025beyond} & - & 32.56 & 79.54 & 80.90 & 75.89 & 87.52 & - \\
BaiChuan-M2-32B~\cite{dou2025baichuan} & 16.78 & 22.80 & 75.05 & 79.20 & 68.49 & 84.68 & 78.73 \\
QuarkMed-32B~\cite{li2025quarkmed} & - & 28.60 & - & 79.00 & 75.50 & 86.02 & 88.61 \\
\midrule
\multicolumn{8}{c}{\textbf{DeepResearch Models}} \\
\midrule
\addlinespace[0.25em]
Tongyi-DeepResearch~\cite{team2025tongyi}-30BA3B & \textbf{28.19} & 35.40 & 79.54 & 76.20 & 75.21 & 87.90 & 87.59 \\
\addlinespace[0.25em]
\rowcolor{cyan!2} \textbf{\textsc{DeepMed}-14B-SFT} (ours) & 19.46 & 34.80 & 77.92 & 80.40 & 79.37 & 87.27 & 90.31 \\
\rowcolor{cyan!2} \textbf{\textsc{DeepMed}-14B-RL} (ours) & 26.84 & \textbf{36.14} & \textbf{79.93} & \textbf{81.80} & \textbf{82.60} & \textbf{88.22} & \textbf{91.19} \\
\bottomrule
\bottomrule
\end{tabular}
}
\caption{Performance comparison of \textsc{DeepMed} and baselines on medical benchmarks ("-" indicates not reported).}
\label{tab:med_benchmarks}
\vspace{-1em}
\end{table*}

\begin{figure}[htbp]
  \centering
  \resizebox{0.43\textwidth}{!}{%
    \includegraphics{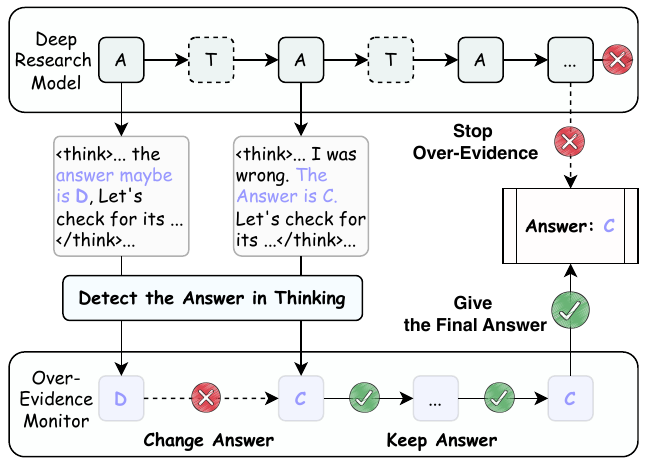}%
  }
  \caption{The workflow of `Over-Evidence' monitor.}
  \label{fig:timeline}
  \vspace{-1em}
\end{figure}

\subsection{Inference with Over-Evidence Monitor}
Although med-enhanced agentic RL has substantially reduced the prevalence of ``Over-Evidence'', the model remains prone to this failure mode when confronted with particularly challenging problems. 
To maintain stable inference, we introduce a lightweight sub-agent that monitors OE during inference, as shown in Fig.~\ref{fig:oem}. 
At each step, it reads the DR model’s response and infers its intent. 
Then it caches any proposed answer; if the cached answer remains unchanged in the next turn, the model is likely invoking tools solely for detail verification. 
Limited validation can be beneficial, but repeated validation without any cache updates signals stalled progress and merely injects noise into the context. 
Accordingly, if the cache remains unchanged for more than a preset number of turns, the monitor terminates the interaction and returns the cached answer as the model’s final output. 
This early-stopping fallback is preferable to letting the model continue iterating under an increasingly corrupted context—where redundant tool calls bloat the prompt, exacerbate context rot, and may end only when the turn budget is exhausted. 

\section{Experiments}
\subsection{Settings}
\noindent \textbf{Baselines and Datasets.}
We select a total of 13 models for our baselines spanning three groups: Awesome General Models, Medical reasoning models, and DeepResearch models. 
We compare on seven plain-text datasets, including medical subsets of several general benchmarks. 
Details of them are provided in the Appendix.~\ref{appendix:settings}.

\noindent \textbf{Training Settings.}
Our model is trained based on the Qwen3-14B model.
We use 5,437 multi-hop med-search data for ASFT and 6,304 difficulty-filtered medical data for ARL. 
More settings are located in Appendix.~\ref{appendix:exp_tool}. 

\noindent \textbf{Tools Configuration.}
We equip the model with a \texttt{search} and a \texttt{visit} tool. 
The former uses query-based search, returning multiple webpage URLs and brief descriptions, while the latter is responsible for retrieving the necessary information from the full content of a specific webpage. 
Tool details are in Appendix.~\ref{appendix:exp_tool}. 

\begin{table*}[h]
\small
\centering
\resizebox{\textwidth}{!}{%
\begin{tabular}{l|cc|cc|cc|cc|cc|cc|cc}
\toprule
\multirow{2}{*}{\textbf{Model}} & \multicolumn{2}{c|}{\textbf{HLE-Med}} & \multicolumn{2}{c|}{\textbf{MedXpert}} & \multicolumn{2}{c|}{\textbf{MMLU-Pro}} & \multicolumn{2}{c|}{\textbf{PubMedQA}} & \multicolumn{2}{c|}{\textbf{MedMCQA}} & \multicolumn{2}{c|}{\textbf{MedQA}} & \multicolumn{2}{c}{\textbf{CMExam}} \\
& \#Acc & \#Turn & \#Acc & \#Turn & \#Acc & \#Turn & \#Acc & \#Turn & \#Acc & \#Turn & \#Acc & \#Turn & \#Acc & \#Turn \\
\midrule
\textsc{DeepMed}-SFT-14B              & 19.46 & 10.4 & 34.80 &  8.8 & 77.92 & 5.8 & 80.40 & 2.7 & 79.37 & 4.8 & 87.27 & 4.9 & 90.31 & 5.3 \\
\midrule
\textsc{DeepMed}-SFT-14B w GENERAL        & 18.12 & 12.2 & 32.89 & 12.0 & 76.71 & 7.2 & 78.80 & 4.0 & 76.88 & 5.4 & 86.72 & 5.8 & 88.69 & 6.5 \\
\textsc{DeepMed}-RL-14B w/o Penalty   & 26.84 & 14.6 & 35.52 & 14.7 & 78.63 & 9.2 & 79.80 & 5.4 & 79.44 & 6.7 & 87.88 & 6.7 & 90.89 & 7.3 \\
\textsc{DeepMed}-RL-14B w/o monitor   & 24.83 & 12.9 & 35.72 & 11.2 & 77.61 & 8.2 & 81.20 & 3.4 & 80.11 & 7.8 & 86.88 & 6.4 & 90.01 & 6.4 \\
\midrule
\textsc{DeepMed}-RL-14B               & \textbf{26.84} & 11.8 & \textbf{36.14} &  9.4 & \textbf{79.93} & 6.2 & \textbf{81.80} & 2.8 & \textbf{82.60} & 5.2 & \textbf{88.22} & 5.0 & \textbf{91.19} & 4.7 \\
\bottomrule
\end{tabular}%
}
\caption{Performance comparision for ablation results: w/o Monitor removes the OE Monitor; w/ General uses general training data; w/o Penalty removes the turn-penalty.}
\label{tab:medical_ablation}
\vspace{-1em}
\end{table*}

\subsection{Main Results}
Tab.~\ref{tab:med_benchmarks} presents the performance comparison between \textsc{DeepMed} and other models on seven medical benchmarks.
We can summarize three insights:
\textbf{(ii)} After Warm-up SFT, \textsc{DeepMed}'s performance on many benchmarks has surpassed most medical expert models with more parameters, especially the more difficult HLE-Med and MedXpert. On some common medical benchmarks, \textsc{DeepMed} even surpassed Tongyi-DR. This demonstrates that pure medical multi-hop problems significantly enhance the model's ability in medical diagnosis.
\textbf{(ii)} After the RL stage, performance improves across all benchmarks. DeepMed achieves state-of-the-art results on every benchmark except HLE-Med. Notably, it does so with fewer parameters and substantially less medical training data, underscoring the strong potential of deepresearch capabilities for medical diagnosis..
\textbf{(iii)} General-purpose Tongyi-DR and strong medical LLMs like AlphaMed and QuarkMed each outperform the other on different benchmarks, suggesting that search capability and medical reasoning offer complementary strengths. 
\textsc{DeepMed} is the bridge of them. 

\subsection{Ablation Study}
In ablation experiments, we mainly need to address the following three questions: 
(i) the quality of our synthesized multi-hop med-search data; 
(ii) the benefits of difficult-aware turn-penalty in ARL; and 
(iii) whether the OE Monitor improves performance. 
So we have three variants: (i) a model using open-source general multi-hop data from Deepdive~\cite{lu2025deepdive} for SFT; (ii) an RL model without turn-penalty; and (iii) removing the OE Monitor from the RL model. 
Results in Tab.~\ref{tab:medical_ablation} answers the question above: removing any component degrades \textsc{DeepMed}’s performance, indicating that each of our proposed improvements contributes meaningfully to final results. 

\begin{figure}[htbp]
  \centering
  \resizebox{0.4\textwidth}{!}{%
    \includegraphics{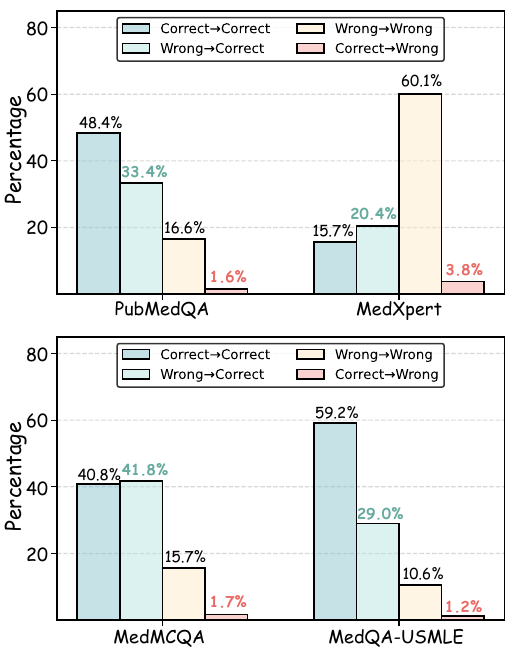}%
  }
  \caption{Answer outcome breakdown: most correct predictions stem from tool use and self-correction, while errors caused by later tool use or reasoning are rare.}
  \label{fig:correct_wrong}
  \vspace{-1em}
\end{figure}
\subsection{Medical Evidence-Seeking Ability Analyse}
In our experiments, we find that under the deepresearch paradigm, \textsc{DeepMed} typically first proposes a tentative answer and then uses search to verify its details, revise the hypothesis, and repeat this search–reflection loop until it converges on a final decision.
To quantify the effectiveness of this search-and-reflection process, we analyze how the model’s initial answer differs from its final answer. Specifically, we categorize each instance into four cases: (i) initially correct and finally correct; (ii) initially incorrect but corrected during search, ending correct; (iii) incorrect throughout; and (iv) initially correct but revised to an incorrect final answer.

As shown in the Fig.~\ref{fig:correct_wrong}, across all four evaluation groups, a large fraction of initially incorrect answers are corrected to become correct, while cases that flip from initially correct to finally incorrect do occur but are much rarer. These results suggest that \textsc{DeepMed} can effectively exhibit an evidence-based pattern of medical reasoning, using retrieved evidence to revise erroneous hypotheses. 
An illustrative case is in Appendix~\ref{appendix:deepmed_solve}.

\begin{figure}[htbp]
  \centering
  \resizebox{\columnwidth}{!}{%
    \includegraphics{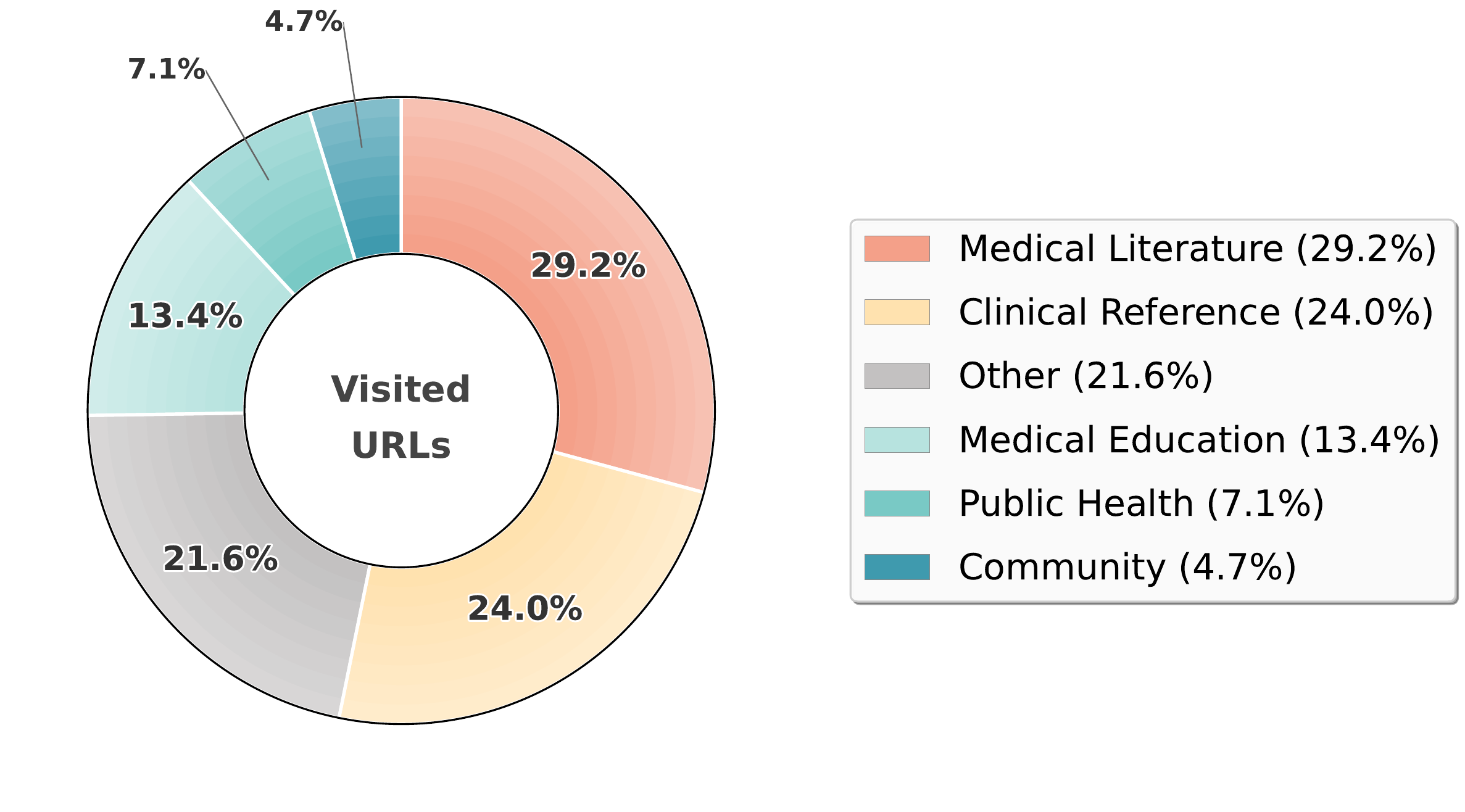}%
  }
  \caption{Distribution of website categories visited by DeepMed via the \texttt{Visit} tool.}
  \label{fig:visit_pie}
  \vspace{-0.5em}
\end{figure}
\begin{table}[htbp]
\centering
\small
\renewcommand{\arraystretch}{1}
\begin{tabular}{ll}
\toprule
\textbf{Category} & \textbf{Representative Root Domains} \\
\midrule
Medical Literature 
& \makecell[l]{\texttt{pubmed.ncbi.nlm.nih.gov} \\ \texttt{nejm.org}} \\

Clinical Reference 
& \makecell[l]{\texttt{uptodate.com} \\ \texttt{mayoclinic.org}} \\

Medical Education 
& \makecell[l]{\texttt{amboss.com} \\ \texttt{medbullets.com}} \\

Public Health 
& \makecell[l]{\texttt{cdc.gov} \\ \texttt{nih.gov}} \\

Medical Community 
& \makecell[l]{\texttt{researchgate.net} \\ \texttt{forums.studentdoctor.net}} \\
\bottomrule
\end{tabular}
\caption{Representative root domains for each category of visited web sources.}
\label{tab:website}
\vspace{-1em}
\end{table}

\subsection{Information Source Details}
Fig.~\ref{fig:visit_pie} shows the distribution of website categories accessed via \texttt{Visit} during training, and Tab.~\ref{tab:website} lists representative root domains. Most visits come from authoritative sources, dominated by Medical Literature (29.2\%; e.g., PubMed/PMC/NEJM) and Clinical References (24.0\%; e.g., UpToDate/Mayo Clinic/Merck Manuals). Medical Education sites account for 13.4\% (e.g., AMBOSS, MedBullets), while Public Health (7.1\%) and Medical Community platforms (4.7\%) play supplementary roles. Notably, even within the \emph{Other} category, a large fraction of visits still come from high-quality knowledge bases such as Wikipedia. Overall, the distribution indicates that \textsc{DeepMed} primarily relies on expert-curated evidence, supporting the reliability of retrieved web knowledge.

\begin{figure}[htbp]
  \centering
  \resizebox{0.47\textwidth}{!}{%
    \includegraphics{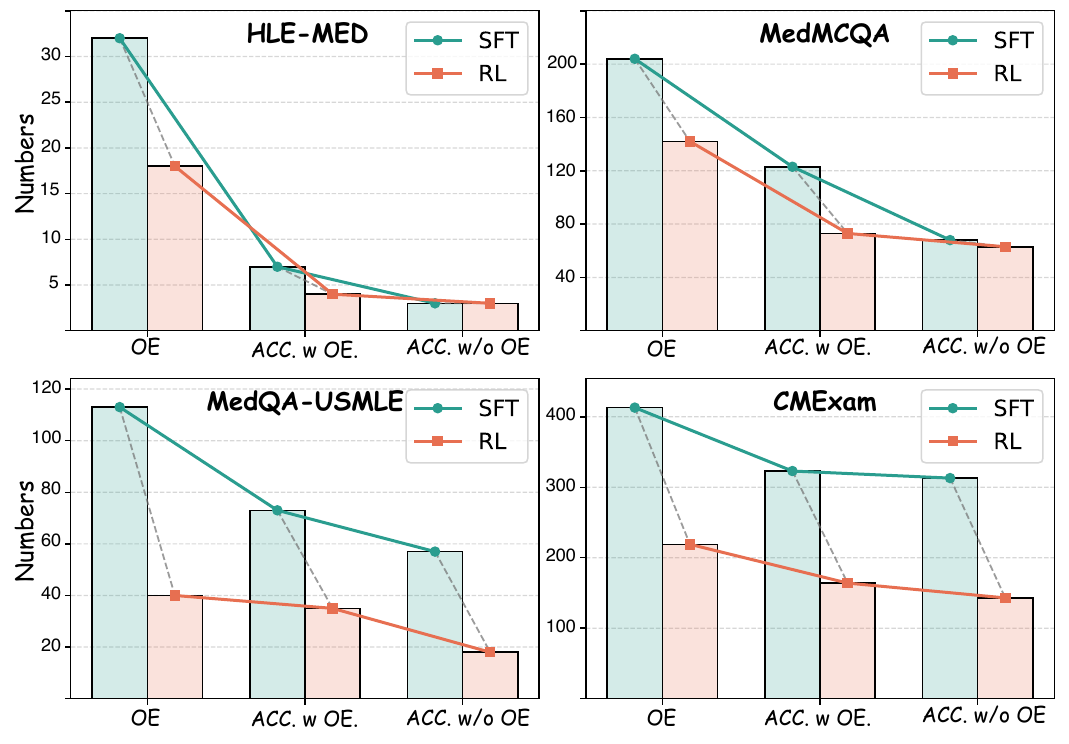}%
  }
  \caption{An illustration of \textsc{DeepMed}-SFT/RL on four datasets: OE Monitor (OEM) trigger counts and the number of correct predictions with OEM versus correct predictions continued inference without OEM. }
  \label{fig:oem}
  \vspace{-0.5em}
\end{figure}
\subsection{Over-Evidence Monitor Analysis}
Fig.~\ref{fig:oem} reports, across four datasets, the numbers of inference runs that enter an over-evidence state, and measure how many of these cases are answered correctly via the Over-Evidence Monitor (OEM) as well as how many remain correct if the model continues reasoning.
We find that over-evidence triggers occur more frequently after the SFT stage, but drop substantially after reinforcement learning (RL). This suggests that training on diagnosis-oriented medical data reduces excessive searching: the model is more likely to stop after gathering key evidence and rely on reasoning to produce an answer. Moreover, once OEM is triggered, the number of correct cases remains consistently low, indicating that when the model enters the over-evidence state it often becomes trapped in an incorrect medical reasoning path and repeatedly “verifies” the same mistaken hypothesis.
On HLE-Med (149 questions), the trigger counts are 32/149 and 18/149, respectively, further showing that harder questions are more likely to drive the model into an over-evidence state.

\begin{figure}[htbp]
  \centering
  \resizebox{0.43\textwidth}{!}{%
    \includegraphics{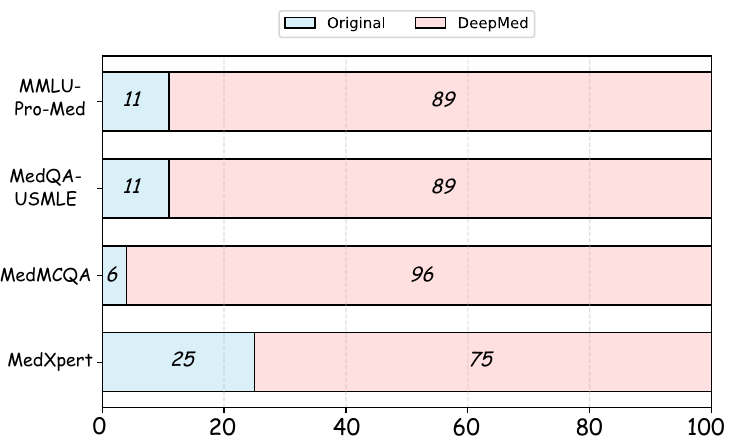}%
  }
  \caption{LLM-based comparison of whether the evidence used by the base model or \textsc{DeepMed} for the final decision is more complete and reliable.}
  \label{fig:against}
  \vspace{-1em}
\end{figure}
\subsection{Evidence and Logical Completeness Analysis}
To more directly assess how \textsc{DeepMed} benefits from the DeepResearch paradigm—particularly in evidence seeking and hallucination reduction—we compare it with its base model on instances where both models answer correctly (MedXpert, MMLU-Pro, MedMCQA, and MedQA-USMLE). 
For each such instance, we extract the evidence chains produced by the two models and use an LLM judge to determine which chain is more comprehensive and logically reliable. 
The evaluation prompt is provided in Appendix.~\ref{appendix:evidence}. 
The results in Fig.~\ref{fig:against} show that \textsc{DeepMed} overwhelmingly produces evidence chains with higher completeness and reliability, suggesting that its medical judgments under the DR paradigm are more credible. 
\section{Conclusion}
We present \textsc{DeepMed}, a DeepResearch model tailored for medical questions that bridges task and tool-scaling gaps between general-purpose DR systems and medical reasoning. \textsc{DeepMed} combines web-based multi-hop med-search data synthesis with Agentic SFT/RL to better couple retrieval with clinical reasoning, and introduces a difficulty-aware turn-penalty and an Over-Evidence Monitor to curb over-searching and noisy context. Experiments show consistent gains over the base model on seven medical benchmarks and improvements over several larger or more data-trained baselines. Further analysis shows that tool-augmented search--reflection--verification improves self-correction and evidence grounding, reducing hallucinations and spurious attributions. 

\section*{Limitations}
\textsc{DeepMed} demonstrates the feasibility of combining the DeepResearch paradigm with medical reasoning using a relatively small amount of training data. However, agentic training and tool-augmented rollouts incur non-trivial computational and engineering overhead, and we have not yet systematically scaled our approach to larger backbone models or substantially larger training corpora. In addition, \textsc{DeepMed} is evaluated only on public benchmarks and has not been validated in real clinical workflows or under regulatory standards; it should not be used as a substitute for professional medical advice or clinical decision making. Finally, our tools retrieve evidence primarily from the open web, whose availability and quality may vary over time. We hope to collaborate with medical institutions in future work to incorporate more stable and authoritative information sources and to enable more realistic evaluations.

\section*{Ethics Statement}
This work does not involve human participants, patient records, or clinical studies, and thus no IRB approval is required. We rely on public benchmarks and synthesized data, and we do not collect or distribute any personally identifiable information. For web-based evidence retrieval, we only access publicly available content and report aggregated findings; we encourage compliance with website terms and applicable data policies. The system is a research prototype and must not be used as medical advice or for clinical decision making without appropriate oversight and validation.
\bibliography{acl_latex}

\clearpage
\appendix

\section{Appendix}
\begin{figure}[h]
    \centering
    \begin{paperbox}{Multi-hop Med-search QA}
        \textbf{Question:} \\
        A pediatric malignancy accounting for roughly 50\% of soft tissue sarcomas in children and approximately 400 to 500 annual diagnoses in the United States comprises distinct histological categories. The most prevalent form, representing 60--70\% of cases with a peak incidence in children aged 0--5 years, is typically "fusion-negative" and often driven by RAS pathway mutations. 
        
        Conversely, a more aggressive subtype, constituting 20--30\% of diagnoses with a peak onset between 10 and 25 years, is characterized by the presence of chimeric transcription factors. In approximately 60\% of this aggressive subtype, the pathology is driven by a specific fusion oncoprotein that is 438 amino acids in length and exhibits 10- to 100-fold greater transcriptional activity than its wild-type predecessors. This protein is distinct from a less common variant found in roughly 20\% of cases. 
        
        Identify the specific reciprocal genetic rearrangement---naming the chromosomes and bands involved---that generates this primary 438-amino acid oncogenic driver.
        
        \vspace{0.3cm}
        \hrule
        \vspace{0.3cm}
        
        \textbf{Reasoning Path:} \\
        Childhood Rhabdomyosarcoma $\rightarrow$ Embryonal vs. Alveolar (Aggressive) $\rightarrow$ Alveolar Rhabdomyosarcoma $\rightarrow$ \textit{PAX3::FOXO1} Fusion $\rightarrow$ \textbf{t(2;13)(q35;q14) translocation}
        
        \vspace{0.3cm}
        \textbf{Answer:} \\
        t(2;13)(q35;q14) translocation
    \end{paperbox}
    
    \caption{A case of Multi-hop Med-Search QA}
    
    \label{fig:oncology_case}
\end{figure}
\subsection{A case of Multi-hop Med-Search QA}
\label{appendix:data}
Fig.~\ref{fig:oncology_case} shows an example of our synthesized multi-hop Med-Search QA. Notably, the question contains many numerical details. We intentionally encourage the inclusion of numerical attributes in the synthesis prompt, since numbers such as prevalence rates and medication dosages are common sources of hallucination compared to entity relations and thus typically require verification via search. This design more strongly incentivizes the model to use tools in medical contexts and promotes deeper, evidence-grounded reasoning. 
Such information is rarely obtainable in a comprehensive form from a single medical website, let alone from a knowledge graph. This highlights the advantage of our web-based synthesis method, which aggregates evidence across multiple authoritative sources.

\subsection{Evaluation Setup}
\label{appendix:settings}
\paragraph{Benchmarks.}
We evaluate DeepMed on seven medical benchmarks as Tab.~\ref{tab:med_benchmarks}, we can splite them to (1) challenging dataset: \textbf{HLE-Med} and \textbf{MedXpert}, and (2) standard dataset: \textbf{MMLU-Pro-Med}, \textbf{PubMedQA}, \textbf{MedMCQA}, \textbf{MedQA},\textbf{CMExam}. 
\begin{itemize}
    \item \textbf{HLE-Med}~\citep{phan2025humanitysexam} stands as one of the most challenging medical benchmarks, featuring questions meticulously crafted and validated by human medical experts. We evaluate the model on the \texttt{Biology/Medicine} text subset \footnote{\url{https://huggingface.co/datasets/futurehouse/hle-gold-bio-chem}}with 149 questions. 
    
    \item \textbf{MedXpert}~\citep{medxpert} is a complex benchmark designed to assess expert-level medical reasoning grounded in relevant medical literature. In our experiments, we utilize the \texttt{text} subset that contains 2,450 questions.
    
    \item \textbf{MMLU-Pro-Med}~\citep{wang2024mmluprorobustchallengingmultitask} provides a comprehensive evaluation of medical knowledge and reasoning capabilities. We employ the \texttt{biology} and \texttt{health} subsets, totaling 1535 questions.
    
    \item \textbf{PubMedQA}~\citep{jin-etal-2019-pubmedqa} contains 500 questions to evaluate the model's ability of answering clinical inquiries based on medical articles.

    \item \textbf{MedMCQA}~\citep{medmcqa} includes 4,183 questions designed to address real-world medical entrance exam questions.
    
    \item \textbf{MedQA}~\citep{medqa} consists of questions collected from professional medical examinations. We focus on the USMLE subset, comprising 1,273 questions.
    
    \item \textbf{CMExam}~\citep{liu2023benchmarking} consists of 6,811 questions sourced from the Chinese National Medical Licensing Examination.
    \vspace{-5mm}
\end{itemize}

\begin{table}[h]
\centering
\small
\setlength{\tabcolsep}{4pt}
\resizebox{\columnwidth}{!}{%
\begin{tabularx}{\columnwidth}{l c >{\raggedright\arraybackslash}X}
\toprule
\textbf{Dataset} & \textbf{Size} & \textbf{Type} \\
\midrule
\textbf{HLE-Med}~\citep{phan2025humanitysexam}   & 149 & MCQ + Open-ended \\
\textbf{MedXpert}~\citep{medxpert}  & 2,450  & MCQ \\
\textbf{MMLU-Pro-Med}~\citep{wang2024mmluprorobustchallengingmultitask}  & 1,535 & MCQ \\
\textbf{PubMedQA}~\citep{jin-etal-2019-pubmedqa}  & 500  & MCQ \\
\textbf{MedMCQA}~\citep{medmcqa}   & 4,183  & MCQ \\
\textbf{MedQA-USMLE}~\citep{medqa}     & 1,273  & MCQ \\
\textbf{CMExam}~\citep{liu2023benchmarking}    & 6,811  & MCQ \\
\bottomrule
\end{tabularx}%
}
\caption{Overview of evaluation benchmarks.}
\label{tab:dataset_overview}
\end{table}

\paragraph{Baselines.}
We compare DeepMed against three categories of strong baselines. 
And we avoid comparing against models that rely on private databases or are not fully open-sourced (e.g., MedResearcher-R1~\cite{yu2025medresearcher}), as such comparisons may introduce confounding factors and bias. 
First, we select representative \textbf{General Large Language Models}, including Gemini-2.5-Pro, Kimi-k2-Thinking~\citep{team2025kimi}, DeepSeek-V3.2~\citep{liu2025deepseek}, and the Qwen3 series (Qwen3-32B-Thinking and Qwen3-14B)~\citep{yang2025qwen3}. 
Second, we evaluate state-of-the-art \textbf{Medical LLMs} that utilize diverse post-training and data synthesis strategies: 
HuatuoGPT-o1-70B~\citep{chen2024huatuogpto1medicalcomplexreasoning} pioneers the SFT-then-RL paradigm to achieve SOTA reasoning performance; 
the M1 series focuses on enhancing medical reasoning via test-time scaling; 
MedReason-8B~\citep{wu2025medreason} employs a medical knowledge graph to synthesize high-quality CoT paths; 
AlphaMed-70B~\citep{liu2025beyond} achieves competitive results by analyzing question informativeness and applying pure RL specifically to high-informativeness questions; 
BaiChuan-M2-32B~\citep{dou2025baichuan} introduces a Patient Simulator and Clinical Rubrics Generator to mimic real-world clinical decision-making, subsequently optimized via interactive GRPO; 
and QuarkMed-32B~\citep{li2025quarkmed} enhances medical knowledge coverage by synthesizing vast amounts of data from medical literature and clinical records. 
Thirdly, we include \textbf{Deep Research Models} designed for autonomous information synthesis, such as Tongyi-DeepResearch~\citep{team2025tongyi} and MedResearcher-R1~\citep{yu2025medresearcher}.

\subsection{Experiments Details}
\label{appendix:exp_tool}
\paragraph{Data Synthesis}
To synthesize multi-hop med-search data, we first need to define suitable starting points. 
We initiate random walks from the 50 most common real-world drugs and diseases. 
During the walks, we store every visited entity description and every accessed webpage in a local cache, which helps reduce cost by avoiding redundant retrieval in subsequent visits. 
In total, we cached 67,024 entity descriptions and 200,584 webpages. When generating multi-hop questions, we randomly sample walk trajectories with hop lengths between 3 and 8, ensuring a more balanced distribution of question difficulty. 
\paragraph{Training Settings}
Both agentic SFT and RL training are conducted using the Verl ~\cite{verl} framework. The training prompt template is shown in the Fig.~\ref{fig:sp} During the agentic SFT stage, we train the model for 5 epochs and use a training batch size of 32 with a learning rate of $1\times10^{-5}$, applying a cosine learning rate schedule with a warm-up ratio of 0.1. The maximum sequence length is set to 40K tokens. For the agentic RL stage, we sample $G=8$ rollouts per query, with a maximum of 30 tool calling turns per rollout. The training batch size is 64, with a PPO mini-batch size of 16. The maximum sequence length remains 40K tokens, the clipping range $\epsilon$ is set to 0.2, and the learning rate is set to $1\times10^{-6}$. 
\paragraph{Inference Settings}
During inference, we set the temperature to 1.0 and used default values for all other parameters. We capped the maximum number of rounds at 30, and limited the OE monitor to 20 rounds. 
\begin{figure}[t]
    \centering
    \begin{paperbox}{Prompt for Evidence Analysis}
        You are an expert evaluator for medical question answering systems. Your task is to judge which model provides \textbf{MORE COMPLETE EVIDENCE} to support its answer.

        \vspace{0.2cm}
        \textbf{Question} \\
        \{question\}

        \vspace{0.2cm}
        \textbf{Model A (Original Model) -- Single-turn reasoning} \\
        \{original\_evidence\}

        \vspace{0.2cm}
        \textbf{Model B (RL Model) -- Multi-turn with web search} \\
        \{rl\_evidence\}

        \vspace{0.2cm}
        \textbf{Evaluation Criteria} \\
        Consider the following aspects when judging evidence completeness:
        \begin{enumerate}[leftmargin=*, label=\arabic*., nosep]
            \item \textbf{Breadth of Sources}: Does the model cite multiple authoritative sources (medical textbooks, clinical guidelines, research papers)?
            \item \textbf{Depth of Explanation}: Does the model explain the underlying medical mechanisms, not just state facts?
            \item \textbf{Clinical Relevance}: Does the model provide clinically relevant details (dosages, contraindications, diagnostic criteria)?
            \item \textbf{Verification}: Does the model verify its claims with external evidence?
            \item \textbf{Reasoning Chain}: Is the logical chain from evidence to conclusion clear and complete?
        \end{enumerate}

        \vspace{0.2cm}
        \textbf{Instructions}
        \begin{itemize}[leftmargin=*, nosep]
            \item Model A uses internal knowledge with detailed reasoning in a single turn.
            \item Model B uses web search to find and cite external sources across multiple turns.
            \item Judge based on \textbf{EVIDENCE COMPLETENESS}, not just answer correctness (both got the correct answer).
            \item Consider both quality and quantity of supporting evidence.
        \end{itemize}

        \vspace{0.2cm}
        \textbf{Response Format} \\
        Respond with a JSON object: \\
        \{ "winner": "A" or "B", "reason": "Brief explanation (1-2 sentences)" \} \\
        Only output the JSON, nothing else.
    \end{paperbox}
    \vspace{-3mm}
    \caption{Prompt for evidence Analysis.}
    \label{fig:eval_prompt}
\end{figure}

\begin{figure}[h]
    \centering
    \begin{tcolorbox}[
        enhanced,
        title=\textbf{Function Calling Schema},
        colback=white,
        colframe=black!90,
        coltitle=white,
        fonttitle=\bfseries\small,
        boxrule=0.8pt,
        arc=2mm,
        attach boxed title to top left={xshift=4mm, yshift=-3mm},
        boxed title style={colback=black!90, sharp corners=south}
    ]
\begin{lstlisting}[language=json]
// Tool 1: Search Tool
{
    "type": "function",
    "function": {
      "name": "search",
      "description": "Perform Google web searches and return top results.",
      "parameters": {
        "type": "object",
        "properties": {
          "query": {
            "type": "array",
            "items": { "type": "string" },
            "description": "Array of query strings (1-5 queries).",
            "minItems": 1,
            "maxItems": 5
          }
        },
        "required": [ "query" ]
      }
    }
},
// Tool 2: Visit Tool
{
    "type": "function",
    "function": {
      "name": "visit",
      "description": "Visit webpage(s) and return summary.",
      "parameters": {
        "type": "object",
        "properties": {
          "url": {
            "type": "array",
            "items": { "type": "string" },
            "description": "The URL(s) to visit (1-3 URLs).",
            "minItems": 1,
            "maxItems": 3
          },
          "goal": { "type": "string" }
        },
        "required": [ "url" ]
      }
    }
}
    \end{lstlisting}
    \end{tcolorbox}
    \vspace{-3mm}
    \caption{JSON schema of the tools used.}
    \label{fig:tool_prompts}
\end{figure}

\paragraph{Tool Configuration}
In this work, we predefine two tools to support medical reasoning: a \texttt{search} tool and a \texttt{visit} tool. The \texttt{search} tool is implemented using Serper’s Google Search API\footnote{\url{https://serper.dev}}, which retrieves web search results based on specified queries and returns both brief summaries and corresponding URLs. 
Jina API\footnote{\url{https://jina.ai}} is used for \texttt{visit} tool to get all webpage content of the specific url. 
Subsequently, we segment the webpage content into chunks of 100k tokens. 
A summarization model then processes these chunks to extract only the information relevant to the given goal, thereby preventing context overflow caused by excessively long webpages. The detailed tool schemas are provided in Fig.~\ref{fig:tool_prompts}. The prompt for this summarization model is shown in Fig.~\ref{fig:summary_model}. 

\begin{figure}
    \centering
\begin{paperbox}{System Prompt}
You are a Medical deep research assistant. Your core function is to conduct thorough, multi-source investigations into any topic. You must handle both broad, open-domain inquiries and queries within specialized academic fields. For every request, synthesize information from credible, diverse sources to deliver a comprehensive, accurate, and objective response. When you have gathered sufficient information and are ready to provide the definitive response, you must enclose the entire final answer within <answer></answer> tags.
\end{paperbox}
\vspace{-3mm}
\caption{System prompt for training.}
\label{fig:sp}
\end{figure}

\begin{figure}[h]
    \centering
    \begin{paperbox}{Prompt for summary model}
        Please process the following webpage content and user goal to extract relevant information:
        
        \vspace{0.3cm}
        \textbf{\large Webpage Content} \\
        \{content\} 
        
        \vspace{0.3cm}
        \textbf{\large User Goal} \\
        \{goal\}    
        
        \vspace{0.3cm}
        \textbf{\large Task Guidelines}
        \begin{enumerate}[leftmargin=*, label=\arabic*.]
            \item \textbf{Content Scanning for Rationale}: Locate the \textbf{specific sections/data} directly related to the user's goal within the webpage content.
            \item \textbf{Key Extraction for Evidence}: Identify and extract the \textbf{most relevant information}. Ensure you do not miss any key details. Output the \textbf{full original context} as much as possible (it can exceed three paragraphs).
            \item \textbf{Summary Output for Summary}: Organize into a concise paragraph with logical flow, prioritizing clarity, and assess the contribution of the information to the goal.
        \end{enumerate}
        \vspace{0.2cm}
        \textbf{ Output Format}: JSON format containing "rational", "evidence", and "summary" fields.
    \end{paperbox}
    
    \caption{Prompt used for the summary model. }
    \label{fig:summary_model}
\end{figure}

\subsection{Prompt for Evidence Analysis}
\label{appendix:evidence}
We use the prompt in Fig.~\ref{fig:eval_prompt} directly compare \textsc{DeepMed} and Qwen-14B to determine which produces more well-supported and reliable answers. 
\subsection{A Case of \textsc{DeepMed} Solving QA}
\label{appendix:deepmed_solve}
Figure~\ref{fig:case_study} presents a case study demonstrating how the model leverages search results to rectify prior biases and enhance medical reasoning. In this case, the model initially exhibited a strong prior bias towards the pancreatic duct (Option B) due to anatomical proximity. However, by iteratively refining its search queries to specifically investigate complications of posterior duodenal ulcers, the model retrieved critical evidence identifying the gastroduodenal artery as the structure most susceptible to erosion. This external verification allowed the model to self-correct and confidently converge on the correct diagnosis (Option D).

\begin{figure*}[p]
\centering
\resizebox{0.9\textwidth}{!}{
    \begin{tcolorbox}[
        enhanced,
        width=\textwidth,
        colback=bgmain,        
        colframe=black!90,
        title=\textbf{Case Study: Self-Correction from Search Results},
        fonttitle=\ttfamily,   
        fontupper=\ttfamily,   
        boxrule=0.5pt,
        arc=2mm,
        left=2mm, right=2mm, top=2mm, bottom=2mm
    ]

    \textbf{User:} A 55-year-old man comes to the physician because of a 3-week history of intermittent burning epigastric pain. His pain improves with antacid use and eating, but returns approximately 2 hours following meals. He has a history of chronic osteoarthritis and takes ibuprofen daily. Upper endoscopy shows a deep ulcer located \textbf{on the posterior wall of the duodenal bulb}. This ulcer is most likely to erode into which of the following structures? \textit{(A) Splenic vein (B) Pancreatic duct (C) Descending aorta (D) Gastroduodenal artery (E) Transverse colon}

    \vspace{0.2cm}
    \hrule
    \vspace{0.2cm}

    \textbf{<think>} \\ We have a medical question: a duodenal ulcer on the posterior wall of the duodenal bulb. Which structure is most likely to be eroded? We need to recall ... The pancreatic duct runs in the posterior wall of the duodenum. \textbf{So the most likely answer is B: pancreatic duct.} But let's verify with resources. Use search.\\ \textbf{</think>} 
    \textcolor{annotred}{\textit{\textbf{[Comment]} The model starts with a strong prior bias towards Option B due to anatomical proximity, ignoring the vascular option initially.}}

    \textbf{<tool\_call>} \\
    {\footnotesize \{ "name": "search", "arguments": \{ "query": ["posterior wall duodenal ulcer most likely to erode into pancreatic duct", "duodenal ulcer complication pancreatic duct erosion"] \} \}}\\
    \textbf{</tool\_call>} 
    
    \textbf{<tool\_response>} \\
    {\footnotesize 
    ... \\
    (Result 9) Title: Hepatic arterial haemorrhage caused by duodenal ulcer Abstract: by S Caragher · 2022  Most posterior wall duodenal bulb ulcers erode into the  \textbf{gastroduodenal artery }... url: https://pmc.ncbi.nlm.nih.gov/articles/PMC9628537/.\\}
    \textbf{</tool\_response>}

    \hdashrule{15.5cm}{0.5pt}{1mm}  \\
    \textcolor{annotred}{\textit{\textbf{[Comment]}  The model is verifying the answer... (\textbf{13 turns)}}} \\
    \hdashrule{15.5cm}{0.5pt}{1mm} 

    \textbf{<think>} \\ The search results suggest that erosion into the \textbf{gastroduodenal artery} is a major complication of posterior ulcers, more so than direct pancreatic duct erosion. I need to confirm the anatomical position of the GDA relative to the posterior duodenal wall to be certain.\\ \textbf{</think>}
    \textcolor{annotred}{\textit{\textbf{[Comment]} The model realized that it might be wrong and further explored the details.}} \\
    \textbf{<tool\_call>}\\
    {\footnotesize \{ "name": "visit", "arguments": \{ "goal": "Find mention of specific structures eroded by posterior duodenal ulcer.", "url": ["https://radiopaedia.org/cases/upper-gi-bleeding-duodenal-ulcer"] \} \}}
    \textbf{</tool\_call>}

    \textbf{<tool\_response>}\\
    {\footnotesize 
    ... Case Discussion: The \textbf{gastroduodenal artery} is located just behind the posterior wall of the first part of the duodenum. Consequently, ulcers in this location may erode into the artery causing severe haemorrhage ... \\}
    \textbf{<tool\_response>}

    \vspace{0.2cm}
    \hrule
    \vspace{0.2cm}
    \textbf{<think>} \\ The duodenal bulb (first part) is closely related to the \textbf{gastroduodenal artery} posteriorly. A deep ulcer on the posterior wall is classically known to erode into this artery, causing life-threatening hemorrhage. While the pancreas is also posterior, the GDA is the structure most susceptible to high-pressure erosion in this specific context. Thus answer: D. We need to call the finish tool.\\ \textbf{</think>}\\
    \textbf{<Answer>} (D) Gastroduodenal artery \textbf{</Answer>} 
    \end{tcolorbox}

}
\caption{\textbf{Example of search-assisted medical reasoning.} The figure shows the model's process of correcting its initial incorrect intuition (Option B) by retrieving external evidence regarding the gastroduodenal artery, leading to the correct diagnosis (Option D).}
\label{fig:case_study}
\end{figure*}

\end{document}